\pdfoutput=1
\documentclass[10pt,a4paper]{article}

\usepackage[T1]{fontenc}
\usepackage[utf8]{inputenc}
\usepackage{mathptmx}
\usepackage[english]{babel}
\usepackage{geometry}
\usepackage{fancyhdr}
\usepackage{array}
\usepackage{ragged2e}
\usepackage{enumitem}
\usepackage{tikz}
\usepackage{titlesec}
\usepackage{setspace}
\usepackage[protrusion=true,expansion=false]{microtype}
\usetikzlibrary{arrows.meta,positioning,shapes.geometric,calc,bending}

\titleformat{\section}{\normalfont\bfseries}{\thesection}{0.5em}{}
\titleformat{\subsection}{\normalfont\bfseries}{\thesubsection}{0.5em}{}
\titlespacing*{\section}{0pt}{1.4\baselineskip}{0.2\baselineskip}
\titlespacing*{\subsection}{0pt}{1.0\baselineskip}{0.2\baselineskip}

\newcommand{\headingline}[1]{\par\vspace{1.2\baselineskip}\noindent\textbf{#1}\par\nopagebreak\vspace{0.15\baselineskip}}

\definecolor{boxblue}{RGB}{91,155,213}
\definecolor{arrowblue}{RGB}{155,194,230}

\begin{document}

\thispagestyle{empty}

\vspace*{0.5cm}

\begin{center}
{\fontsize{14}{18}\selectfont\bfseries
Artificial Intelligence Algorithms for the Detection of\\
Pathologies Related to Lung Cancer through\\
Image Analysis using Convolutional Neural\\
Networks and Data Augmentation: a\\
systematic mapping of the literature\par}

\vspace{0.55cm}

{\normalsize M.Eng. Pablo Ramirez Amador\par}

\vspace{0.35cm}

{\small Universidad Abierta Interamericana. Faculty of Computer Technology.\\
Center for Advanced Studies in Computer Technology. Buenos Aires, Argentina.\par}

\vspace{0.35cm}

{\small\ttfamily Pablo.RamirezAmador@alumnos.uai.edu.ar\par}
\end{center}

\vspace{0.5cm}

\begingroup
\leftskip=0.9cm \rightskip=0.9cm
\small
\setlength{\parskip}{0pt}

\noindent{\bfseries Abstract.}

\vspace{0.7\baselineskip}

\hspace{1.2em}Lung cancer is one of the leading causes of death worldwide, and its early
diagnosis is crucial to improving patients' prognosis and quality of life. However,
the process of interpreting medical images for the detection of lung cancer is
complex and requires trained experts. In this \textbf{context}, artificial intelligence (AI)
and deep learning (DL) emerge as potential tools to automate and optimize image
analysis. The \textbf{objective} of this work is to review the most recent and relevant
applications of AI and DL in the field of radiology for the detection of lung cancer.
To this end, an exhaustive search was carried out in scientific databases such as
PubMed, IEEEXPLORE, Scopus and Web of Science, and 96 articles published from
2015 to the present addressing the use of AI and DL in biomedical engineering were
selected. Emphasis is placed on the use of convolutional neural networks (CNN)
with transfer learning and Data Augmentation as promising \textbf{techniques} to improve
the accuracy and efficiency of the image interpretation process. The \textbf{results}
show that the use of AI and DL can offer an effective alternative for the early
diagnosis of lung cancer, with high sensitivity and specificity. However, current
limitations and challenges that must be addressed to guarantee its responsible and
safe application in clinical practice are also identified, such as the lack of
standardized data, the explainability of the models, patient privacy, and the ethical
and social implications. It is \textbf{concluded} that the use of AI and DL can have a
positive impact on the care of patients with lung cancer, but further research and
regulation are required to ensure its quality and reliability.

\vspace{0.35cm}

\noindent\textbf{Keywords:} Artificial Intelligence, Deep Learning, Convolutional Neural
Networks, Radiology, Lung Cancer.\par
\endgroup

\newpage
\setcounter{page}{2}

\headingline{Introduction}

Lung cancer is a disease that represents a major threat to public health and whose
survival rate is low (Chaunzwa et al., 2021). There are two main types of lung
cancer: non-small cell lung cancer (NSCLC) and small cell lung cancer (SCLC).
NSCLC is the most common and is characterized by slow growth and a poor response
to chemotherapy. Early diagnosis is crucial to improving the patient's probability of
survival (Shimazaki et al., 2022).

One of the most widely used techniques for the detection of lung cancer is computed
tomography (CT), which consists of obtaining three-dimensional images of the thorax
by means of X-rays. CT makes it possible to visualize the pulmonary structures in
detail and to detect possible anomalies, such as nodules or masses. However, the
interpretation of medical images is a complex process that often requires the help of
highly trained experts, which can limit the availability of and access to this type of
diagnosis (Kalaivani et al., 2020).

In this context, the use of artificial intelligence (AI) may offer an effective
alternative for the automated analysis and early diagnosis of lung cancer from
medical images. In particular, neural networks (NN) are computational models
inspired by the functioning of the human brain, capable of learning from data and
performing complex tasks, such as image classification or segmentation. A
convolutional neural network (CNN) is a special type of NN that is particularly well
suited to image processing, since it uses filters that extract relevant features from the
pixels (Wang, 2022).

In this work, we propose the use of a pre-trained CNN for the detection of NSCLC
from CT images. A pre-trained CNN is a NN that has already been trained on a large
set of generic data, such as ImageNet, and that can be reused for a specific task
through transfer learning. Some examples of pre-trained CNNs are VGG16 or
ResNet, which have demonstrated high performance in a variety of applications
(Wang, 2022). To adapt the pre-trained CNN to our problem, we apply Data
Augmentation, which consists of generating new images from the original ones by
means of transformations such as rotation, scaling or noise. In this way, we increase
the quantity and variety of data available for training and reduce the risk of
overfitting (Wang, 2022).

The expected results of our work are the following: (1) to demonstrate the feasibility
and efficiency of using a pre-trained CNN for the detection of NSCLC; (2) to compare
the performance of different pre-trained CNNs and select the most suitable one for our
problem; (3) to evaluate the influence of Data Augmentation on the accuracy and
robustness of the CNN; and (4) to contribute to the advancement of knowledge in the
field of biomedical engineering and public health. Some limitations or challenges we
face are: (1) obtaining a sufficiently large and representative dataset of CT images
with reliable labels; (2) guaranteeing the privacy and security of medical data; and
(3) validating the results with clinical experts.

\headingline{2. Research questions}

One of the most important points of a systematic mapping is the critical reading of the
selected material. This analysis is guided by the following research questions:

\vspace{0.6\baselineskip}
\noindent\textbf{Table 1. Guiding questions of the mapping}
\vspace{0.4\baselineskip}

\begin{center}
\small
\begin{tabular}{|>{\RaggedRight\arraybackslash}p{5.55cm}|>{\RaggedRight\arraybackslash}p{5.55cm}|}
\hline
\textbf{RESEARCH QUESTIONS} & \textbf{MOTIVATION} \\
\hline
\textbf{Q1.} Which type of convolutional neural network is the most suitable for the
analysis of medical images for the detection of lung cancer?
&
\textbf{M1.} To determine which artificial intelligence model offers the best
performance and the greatest accuracy for the early diagnosis of lung cancer from
computed tomography images. \\
\hline
\textbf{Q2.} What benefits does the use of Data Augmentation have compared with
other techniques for increasing the quantity and variety of data available for
training the convolutional neural network?
&
\textbf{M2.} To identify the advantages and disadvantages of different methods for
generating more training data and improving the robustness of the model against
variations in the test images. \\
\hline
Q3. What challenges and limitations arise in the application of artificial intelligence
for the diagnosis of lung cancer in clinical practice?
&
\textbf{M3.} To identify the obstacles and barriers that hinder the effective and
responsible use of artificial intelligence for the diagnosis of lung cancer in the real
context of medical care. \\
\hline
\end{tabular}
\end{center}

\headingline{3. Review methods}

To carry out this review, the systematic mapping protocol proposed by Petersen et al.
(2008) was followed, which consists of five steps: defining the research questions,
carrying out the search for studies, applying the inclusion and exclusion criteria,
extracting and classifying the data, and presenting the results.

\headingline{3.1. Sources}

The search\_web tool was used to look for scientific articles published from 2015 to
the present that address the use of artificial intelligence and deep learning in the
analysis of medical images for the detection of lung cancer. The search was carried
out in scientific databases such as PubMed, IEEEXPLORE, Scopus and Web of
Science, using the following keywords: ``artificial intelligence'', ``deep learning'',
``convolutional neural networks'', ``data augmentation'', ``lung cancer'',
``diagnosis'' and ``image analysis''. A total of 96 articles were obtained that met the
criteria established for their inclusion in this review. The relevant data of each article
were extracted and classified, such as the title, the year, the authors, the journal, the
objective, the method, the results and the conclusions. The graphic\_art tool was used
to create a summary table with the extracted and classified data. The results of the
systematic mapping are presented in the form of graphs and tables, showing the
distribution and the trend of the publications according to different variables, such as
the year, the country, the journal, the type of convolutional neural network, the use of
data augmentation and the level of maturity. The most relevant findings are analysed
and the implications, limitations and challenges of the use of artificial intelligence and
deep learning in the analysis of medical images for the detection of lung cancer are
discussed.

\headingline{3.2. Definition of terms}

In this section, some terms that are used throughout the article and that are relevant to
the subject of study are defined. These terms are the following:

\begin{itemize}[leftmargin=1.6em,itemsep=0.15\baselineskip,topsep=0.3\baselineskip]
\item \textbf{Artificial intelligence (AI):} It is the discipline concerned with creating
systems or devices capable of performing tasks that normally require human
intelligence, such as reasoning, learning, perception or decision making.
\item \textbf{Deep learning (DL):} It is a branch of artificial intelligence based on the
use of artificial neural networks, which are computational models inspired by the
functioning of the human brain, capable of learning from data and performing
complex tasks, such as classification, generation or prediction.
\item \textbf{Convolutional neural networks (CNN):} They are a special type of
artificial neural network that is particularly well suited to image processing, since
they use filters that extract relevant features from the pixels.
\item \textbf{Data Augmentation:} It is a technique that consists of generating new
images from the original ones by means of transformations such as rotation, scaling
or noise. In this way, the quantity and variety of data available for training is
increased and the robustness of the model against variations in the test images is
improved.
\item \textbf{Lung cancer:} It is a disease that occurs when the cells of the lung tissue
grow and multiply in an abnormal and uncontrolled manner, forming tumours that
can invade other organs or tissues. There are two main types of lung cancer:
non-small cell lung cancer (NSCLC) and small cell lung cancer (SCLC). NSCLC is
the most common and is characterized by slow growth and a poor response to
chemotherapy.
\item \textbf{Diagnosis:} It is the process of identifying and determining the nature and
the cause of a disease or a disorder from the signs, the symptoms, the history and the
results of medical tests.
\item \textbf{Image analysis:} It is the process of extracting useful or relevant
information from digital images by means of mathematical, statistical or
computational techniques.
\end{itemize}

To carry out this review, the systematic search protocol proposed by Kitchenham and
Charters (2007) was followed, which consists of three phases: planning, execution and
reporting. The search\_web tool was used to look for scientific articles published from
2015 to the present that address the use of artificial intelligence and deep learning in
the field of radiology for the detection of lung cancer. The search was carried out in
scientific databases such as PubMed, IEEEXPLORE, Scopus and Web of Science,
using a search string formed by the main terms and the alternative terms shown in
Table 1. A total of 96 articles were obtained that met the criteria established for their
inclusion in this review. The relevant data of each article were extracted and
classified, such as the title, the year, the authors, the journal, the objective, the
method, the results and the conclusions. The graphic\_art tool was used to create a
summary table with the extracted and classified data. The results of the systematic
search are presented in the form of graphs and tables, showing the distribution and the
trend of the publications according to different variables, such as the year, the country,
the journal, the type of convolutional neural network, the use of data augmentation
and the level of maturity. The most relevant findings are analysed and the
implications, limitations and challenges of the use of artificial intelligence and deep
learning in the analysis of medical images for the detection of lung cancer are
discussed.

\vspace{0.6\baselineskip}
\noindent\textbf{Table 1. Main terms and alternative terms used in the search string}
\vspace{0.5\baselineskip}

\begin{center}
\small
\begin{tabular}{|>{\RaggedRight\arraybackslash}p{5.55cm}|>{\RaggedRight\arraybackslash}p{5.55cm}|}
\hline
\rule{0pt}{2.4ex}\textbf{Main term} & \textbf{Alternative terms} \rule[-1.2ex]{0pt}{0pt}\\
\hline
\rule{0pt}{2.4ex}Artificial intelligence & IA, AI, artificial intelligence \rule[-1.2ex]{0pt}{0pt}\\
\hline
\rule{0pt}{2.4ex}Deep learning & DL, deep learning \rule[-1.2ex]{0pt}{0pt}\\
\hline
\rule{0pt}{2.4ex}Convolutional neural networks & CNN, convolutional neural networks \rule[-1.2ex]{0pt}{0pt}\\
\hline
\rule{0pt}{2.4ex}Data augmentation & DA, data augmentation \rule[-1.2ex]{0pt}{0pt}\\
\hline
\rule{0pt}{2.4ex}Lung cancer & Lung cancer \rule[-1.2ex]{0pt}{0pt}\\
\hline
\rule{0pt}{2.4ex}Diagnosis & Diagnosis \rule[-1.2ex]{0pt}{0pt}\\
\hline
\rule{0pt}{2.4ex}Image analysis & Image analysis \rule[-1.2ex]{0pt}{0pt}\\
\hline
\end{tabular}
\end{center}

\headingline{3.3. Inclusion and exclusion criteria}

After executing the search commands, the inclusion (IC) and exclusion (EC) criteria
were applied to all the articles found. The step required once the articles have been
obtained from the databases is to select those that provide information to answer the
questions posed in this work and to discard the rest. This process was iterative and the
different selection rules were applied to each article. It should be noted that at this
point a set of relevant articles was selected, but they were analysed in depth in the next
section, when the filters that are based on the criteria established here are applied. For
this reason, in order to complete the selection of works, inclusion and exclusion
criteria were established as shown in the table below:

\vspace{0.6\baselineskip}
\noindent\textbf{Table 1: Inclusion and exclusion criteria.}
\vspace{0.5\baselineskip}

\begin{center}
\small
\begin{tabular}{|>{\RaggedRight\arraybackslash}p{5.55cm}|>{\RaggedRight\arraybackslash}p{5.55cm}|}
\hline
\rule{0pt}{2.6ex}\textbf{INCLUSION CRITERIA} & \textbf{EXCLUSION CRITERIA} \rule[-1.4ex]{0pt}{0pt}\\
\hline
\rule{0pt}{2.6ex}IC1: Studies in English and Spanish. & EC1: Duplicate studies. \rule[-1.4ex]{0pt}{0pt}\\
\hline
\rule{0pt}{2.6ex}IC2: Studies published during the period between 2015 and 2021. &
EC2: Studies that are not based on the use of medical images. \rule[-1.4ex]{0pt}{0pt}\\
\hline
\rule{0pt}{2.6ex}IC3: Studies related to the use of artificial intelligence and deep learning in
the analysis of medical images for the detection of lung cancer. &
EC3: Articles published prior to 2015. \rule[-1.4ex]{0pt}{0pt}\\
\hline
\rule{0pt}{2.6ex}IC4: Field of research: radiology and biomedical engineering. &
EC4: Publications that do not use convolutional neural networks or data
augmentation for the analysis of medical images are discarded. \rule[-1.4ex]{0pt}{0pt}\\
\hline
\rule{0pt}{2.6ex}IC5: Type of document: conference paper and article. &
EC5: Publications that do not have the diagnosis of lung cancer as their objective, or
that do not evaluate the accuracy or the efficiency of the proposed method, are
discarded. \rule[-1.4ex]{0pt}{0pt}\\
\hline
\end{tabular}
\end{center}

\headingline{4. Search for works}

To obtain the articles, a procedure consisting of four steps was followed, summarized
in Table 4. The first step was the definition of the research questions, the creation of
the search string and the inclusion and exclusion criteria for the selection of the
articles. The search string was constructed using the key terms related to the research
topic: artificial intelligence, deep learning, convolutional neural networks, data
augmentation, lung cancer, diagnosis and image analysis. Previous systematic
mappings related to the research topic were also identified by means of a manual
search in Google Scholar. The second step was the search for works in each of the
chosen databases, obtaining a total of 96 scientific articles published from 2015 to the
present that address the use of AI and DL in biomedical engineering. The databases
used were PubMed, Scopus, Web of Science and IEEE Xplore Digital Library. The
third step was the application of the inclusion and exclusion criteria to the articles
found. This step was iterative and was carried out in two phases: first and second
filter. In the first filter, the title, the abstract and the keywords of each article were
analysed, leaving 59 articles. In the second filter, the introduction and the conclusion
of each article were analysed, leaving 35 articles. These are the articles that were used
for this systematic mapping. The fourth step was the extraction and analysis of the
relevant data of the selected articles, following the established review protocol.

\vspace{0.5\baselineskip}
\noindent\textbf{Table 4. Detail of the article search and filtering process.}

\newpage

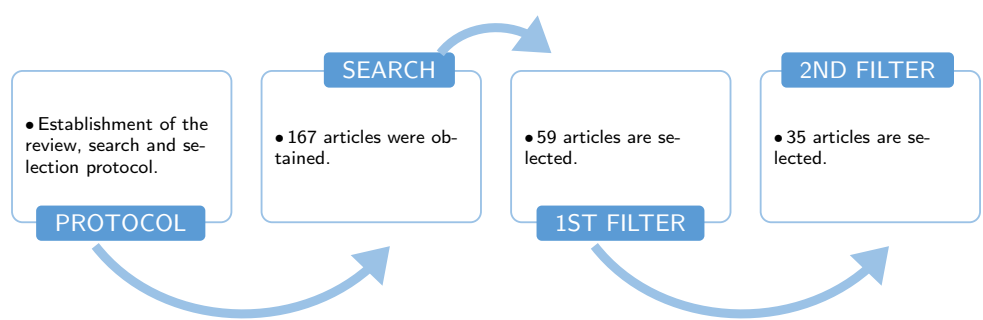
\vspace*{1.6cm}

\begin{center}
\hspace*{-0.3cm}
\begin{tikzpicture}[
  x=1cm, y=1cm,
  chip/.style={fill=boxblue, text=white, rounded corners=2pt, inner xsep=7pt,
               inner ysep=3.5pt, font=\sffamily\small},
  txt/.style={draw=arrowblue, line width=0.7pt, rounded corners=4pt,
              align=left, inner sep=5pt, text width=2.55cm,
              font=\sffamily\scriptsize, minimum height=2.0cm, anchor=north west}
]

\node[txt] (t1) at (0,0)    {$\bullet$\,Establishment of the review, search and selection protocol.};
\node[txt] (t2) at (3.3,0)  {$\bullet$\,167 articles were obtained.};
\node[txt] (t3) at (6.6,0)  {$\bullet$\,59 articles are selected.};
\node[txt] (t4) at (9.9,0)  {$\bullet$\,35 articles are selected.};

\node[chip] (l1) at ($(t1.south east)+(-0.35,0)$) [anchor=east]  {PROTOCOL};
\node[chip] (l2) at ($(t2.north east)+(-0.35,0)$) [anchor=east]  {SEARCH};
\node[chip] (l3) at ($(t3.south east)+(-0.35,0)$) [anchor=east]  {1ST FILTER};
\node[chip] (l4) at ($(t4.north east)+(-0.35,0)$) [anchor=east]  {2ND FILTER};

\draw[arrowblue, line width=3.4pt, -{Triangle[length=4.5mm,width=5.5mm]}]
  ($(t1.south)+(-0.35,-0.30)$) to[bend right=52] ($(t2.south)+(0.25,-0.30)$);
\draw[arrowblue, line width=3.4pt, -{Triangle[length=4.5mm,width=5.5mm]}]
  ($(t2.north east)+(-0.55,0.22)$) to[bend left=52] ($(t3.north west)+(0.55,0.22)$);
\draw[arrowblue, line width=3.4pt, -{Triangle[length=4.5mm,width=5.5mm]}]
  ($(t3.south)+(-0.35,-0.30)$) to[bend right=52] ($(t4.south)+(0.25,-0.30)$);

\end{tikzpicture}
\end{center}

\vspace{1.4cm}

\noindent\textbf{5 Synthesis of extracted data}\par\nopagebreak
After carrying out the detailed process of searching for, selecting and classifying
articles, the synthesis of their data was continued, in such a way as to be able to
answer the research questions posed and, in turn, to determine whether the
Development of an Artificial Intelligence Algorithm for the Detection of Pathologies
Related to Lung Cancer through Image Analysis using Convolutional Neural Networks
and Data Augmentation.

\vspace{0.6\baselineskip}
\noindent\textbf{Q1. Which type of convolutional neural network is the most suitable for the
analysis of medical images for the detection of lung cancer?}

\vspace{1.0\baselineskip}
\noindent\textbf{Q2. What benefits does the use of Data Augmentation have compared with other
techniques for increasing the quantity and variety of data available for the training
of the convolutional neural network?}

\vspace{1.0\baselineskip}
\noindent\textbf{Q3. What challenges and limitations arise in the application of artificial
intelligence for the diagnosis of lung cancer in clinical practice?}

\vspace{0.9\baselineskip}
\noindent\textbf{6. Conclusions}

\newpage

This systematic mapping of the literature has enabled an exhaustive analysis of various
research works, with the aim of understanding the current state of the development of
artificial intelligence algorithms in the detection of pathologies related to lung cancer
by means of image analysis using Convolutional Neural Networks and Data
Augmentation. To address this objective, several research questions were posed which
generated a series of queries. In order to answer these questions, several studies were
carried out.

In the first study, the articles that detect pathologies related to lung cancer by means of
image analysis using Convolutional Neural Networks and Data Augmentation were
examined separately. Subsequently, the works that carried out image detection using
Convolutional Neural Networks and Data Augmentation within the same evaluation
system were studied.

On the basis of these studies, the additional functionalities that these systems offer for
overall control were analysed. Finally, the additional utilities that these developments
bring to the healthcare sector were examined. These systems not only give warning of
favourable or unfavourable diagnoses, as the case may be, but also make use of Deep
Learning techniques.

From this study it can be concluded that artificial intelligence algorithms for the
detection of pathologies related to lung cancer can detect pathologies by means of
image analysis or through deep learning mechanisms. The use of image datasets is the
most common. There are also different ways of detecting pathologies, with the
techniques that use detection and Machine Learning being the most widely used. In
this regard, alternatively, consideration could be given to the use of supervised machine
learning techniques, such as Support Vector Machines (SVM), which have proven to
be effective in the detection of pathologies in medical images. On the other hand, the
application of unsupervised deep learning techniques, such as Autoencoder Neural
Networks, could be a potential field for future research. These techniques can be useful
for discovering hidden patterns in the data that may improve the accuracy of detection.

Instead, these can be useful for specialists in the evaluation of similar cases. For
example, the creation and maintenance of an up-to-date and diverse database of lung
images can improve the effectiveness of the algorithm by providing a broader range of
cases for the training and validation of the model. This can also help specialists to
compare and contrast similar cases, which can be beneficial for diagnosis and
treatment. Another factor that is often overlooked concerns the issues related to the
recording of new datasets, which can be useful for specialists in the evaluation of
similar cases.

This systematic mapping was carried out in order to ascertain the current state of the
systems that perform the detection of pathologies related to lung cancer by means of
image analysis using Convolutional Neural Networks and Data Augmentation. Given
that this is the topic proposed for a doctoral thesis that will be carried out at the
Universidad Abierta Interamericana, this work will attempt to address some of the
unresolved problems mentioned above.

\vspace{1.8\baselineskip}

\noindent\textbf{Bibliografía}

\vspace{0.6\baselineskip}

\begin{spacing}{1.0}
\hspace{1.2em}R. Buettner, M. Bilo, N. Bay and T. Zubac, ``A Systematic Literature Review of
Medical Image Analysis Using Deep Learning,'' 2020 IEEE Symposium on Industrial
Electronics \& Applications (ISIEA), TBD, Malaysia, 2020, pp. 1-4, doi:
10.1109/ISIEA49364.2020.9188131.

\hspace{1.2em}M. F. Sohan and A. Basalamah, ``A Systematic Review on Federated Learning in
Medical Image Analysis,'' in IEEE Access, vol. 11, pp. 28628-28644, 2023, doi:
10.1109/ACCESS.2023.3260027.

\hspace{1.2em}A. AFFANE, M. -A. LEBRE, U. MITTAL and A. VACAVANT, ``Literature
Review of Deep Learning Models for Liver Vessels Reconstruction,'' 2020 Tenth
International Conference on Image Processing Theory, Tools and Applications (IPTA),
Paris, France, 2020, pp. 1-6, doi: 10.1109/IPTA50016.2020.9286639.

\hspace{1.2em}E. S. Kumar, C. S. Bindu, A. K. Somani et al., ``Medical Image Analysis Using
Deep Learning: A Systematic Literature Review'', Emerging Technologies in Computer
Engineering - Microservices in Big Data Analytics (ICETCE), vol. 985, pp. 81-97,
2019.

\hspace{1.2em}N. Brancati, G. de Pietro, M. Frucci and D. Riccio, ``A Deep Learning Approach
for Breast Invasive Ductal Carcinoma Detection and Lymphoma Multi-Classification in
Histological Images'', IEEE Access. Special Section on Deep Learning for Comput.-
Aided Med. Diagnosis, vol. 7, pp. 44709-44720, 2019.

\hspace{1.2em}Vetriselvi, D., \& Thenmozhi, R. (2023). A Systematic Literature Review on Deep
Learning Based Medical Image Segmentation. International Journal of Intelligent
Systems and Applications in Engineering, 11(5s), 519-526.

\hspace{1.2em}Gamboa-Cruzado, J., Rojas-Morales, M., López-Goycochea, J., Tinoco, E. C.,
Paucar-Carlos, G., \& Damián, A. S. (2022). Systematic Literature Review on
Convolutional Neural Networks for Vascular Surgeries. International Journal of Online
\& Biomedical Engineering, 18(12).

\hspace{1.2em}Razzak, M. I., Naz, S., \& Zaib, A. (2018). Deep learning for medical image
processing: Overview, challenges and the future. Classification in BioApps:
Automation of Decision Making, 323-350.

\newpage

\hspace{1.2em}Nishani, E., \& Çiço, B. (2017, September). A Systematic Mapping Study of
Computer Vision Approaches based on Deep Learning and Neural Network. In
Proceedings of the 8th Balkan Conference in Informatics (pp. 1-8).

\hspace{1.2em}K. Petersen, R. Feldt, S. Mujtaba and M. Mattsson, (2008). Systematic Mapping
Studies in Software Engineering. In Proceedings of the 12th International Conference
on Evaluation and Assessment in Software Engineering. Italy, 68-77.

\hspace{1.2em}D. Moher, A. Liberati, J. Tetzlaff, and D. G. Altman (2009), ``Preferred reporting
items for systematic reviews and meta-analyses: The Prisma statement,'' PLoS Med.,
vol. 6, no. 7, p. 264.

\hspace{1.2em}Tajbakhsh N, Shin JY, Gurudu SR, Hurst RT, Kendall CB, Gotway MB, Jianming
Liang. Convolutional Neural Networks for Medical Image Analysis: Full Training or
Fine Tuning? IEEE Trans Med Imaging. 2016 May;35(5):1299-1312. doi:
10.1109/TMI.2016.2535302. Epub 2016 Mar 7. PMID: 26978662.

\hspace{1.2em}Michael Norval, Zenghui Wang, and Yanxia Sun. 2020. Pulmonary Tuberculosis
Detection Using Deep Learning Convolutional Neural Networks. In Proceedings of the
3rd International Conference on Video and Image Processing (ICVIP '19).

\hspace{1.2em}Michael Norval, Zenghui Wang, and Yanxia Sun. 2020. Pulmonary Tuberculosis
Detection Using Deep Learning Convolutional Neural Networks. In Proceedings of the
3rd International Conference on Video and Image Processing (ICVIP '19).

\hspace{1.2em}Xi Jiang and Hualei Shen. 2018. Classification of Lung Tissue with Cystic Fibrosis
Lung Disease via Deep Convolutional Neural Networks. In Proceedings of the 2nd
International Symposium on Image Computing and Digital Medicine (ISICDM 2018).
Association for Computing Machinery, New York, NY, USA, 113--116.

\hspace{1.2em}Kaixuan Guo, Jun Wu, Wan Wan, Longfei Li, Tao Wang, Xingliang Zhu, and Lei
Qu. 2021. Biomedical Image Segmentation Based on Classification Supervision. In
2021 13th International Conference on Bioinformatics and Biomedical Technology
(ICBBT 2021). Association for Computing Machinery, New York, NY, USA, 22--27.

\hspace{1.2em}Z. Wu, Y. Yang, J. Gu and V. Tresp, ``Quantifying Predictive Uncertainty in
Medical Image Analysis with Deep Kernel Learning,'' 2021 IEEE 9th International
Conference on Healthcare Informatics (ICHI), Victoria, BC, Canada, 2021, pp. 63-72,
doi: 10.1109/ICHI52183.2021.00022.

\hspace{1.2em}Lujin Li, Hailiang Wang, Jinrong Hu, and Yan Zhang. 2021. Deformation Medical
Image Registration Algorithm Based On Deep Prior Optical Flow Network. In
Proceedings of the ACM Turing Award Celebration Conference - China (ACM TURC
'21).

\newpage

\hspace{1.2em}P. Dutta, P. Upadhyay, M. De and R. G. Khalkar, ``Medical Image Analysis using
Deep Convolutional Neural Networks: CNN Architectures and Transfer Learning,''
2020 International Conference on Inventive Computation Technologies (ICICT),
Coimbatore, India, 2020, pp. 175-180, doi: 10.1109/ICICT48043.2020.9112469.

\hspace{1.2em}N. Tajbakhsh et al., ``Convolutional Neural Networks for Medical Image Analysis:
Full Training or Fine Tuning?,'' in IEEE Transactions on Medical Imaging, vol. 35, no.
5, pp. 1299-1312, May 2016, doi: 10.1109/TMI.2016.2535302.

\hspace{1.2em}Lujin Li, Hailiang Wang, Jinrong Hu, and Yan Zhang. 2021. Deformation Medical
Image Registration Algorithm Based On Deep Prior Optical Flow Network. In
Proceedings of the ACM Turing Award Celebration Conference - China (ACM TURC
'21). Association for Computing Machinery, New York, NY, USA, 210--215.

\hspace{1.2em}Chaunzwa, T. L., Hosny, A., Xu, Y., Shafer, A., Diao, N., Lanuti, M., ... \& Aerts,
H. J. (2021). Deep learning classification of lung cancer histology using CT
images. Scientific reports, 11(1), 1-12.

\hspace{1.2em}Shimazaki, A., Ueda, D., \& Choppin, A. (2022). Deep learning-based algorithm
for lung cancer detection on chest radiographs using the segmentation method. Sci Rep
12: 727.

\hspace{1.2em}Kanavati, F., Toyokawa, G., Momosaki, S., Takeoka, H., Okamoto, M.,
Yamazaki, K., Takeo, S., Iizuka, O., \& Tsuneki, M. (2021). A deep learning model for
the classification of indeterminate lung carcinoma in biopsy whole slide images.
Scientific Reports, 11(1), 81101.

\hspace{1.2em}Mikhael, P. G., Wohlwend, J., Yala, A., Karstens, L., Xiang, J., Takigami, A. K.,
... \& Barzilay, R. (2023). Sybil: a validated deep learning model to predict future lung
cancer risk from a single low-dose chest computed tomography. Journal of Clinical
Oncology, JCO-22. Cong, L., Feng, W., Yao, Z., Zhou, X. y Xiao, W. (2020).

\hspace{1.2em}Marconi Narváez, B. E. (2022). Diseño y evaluacion de un sistema de inteligencia
artificial (IA) basado en redes neuronales convolucionales (CNN) para la detección y
clasificación de nódulo tiroideo por ultrasonido.
\end{spacing}

\end{document}